\documentclass[12pt]{article}

\usepackage{bm}
\usepackage{amsmath}
\usepackage{amssymb}
\usepackage{graphicx}
\usepackage{esint}
\usepackage{epsf,epsfig,fullpage}
\usepackage{txfonts}
\usepackage{algorithmic}
\usepackage{algorithm}
\usepackage{graphics}
\usepackage{color}
\usepackage{amsfonts}
\usepackage{subfigure}
\usepackage{setspace}
\usepackage{placeins}
\usepackage{float}
\usepackage{textgreek}
\usepackage{authblk}
\usepackage{cite}
\usepackage{titlesec}
\usepackage{hyperref}

\newcommand{\beq}{\begin{equation}}
\newcommand{\eeq}{\end{equation}}
\newcommand{\pder}[2][]{\frac{\partial#1}{\partial#2}}
\newcommand{\pderder}[2][]{\frac{\partial^2#1}{\partial#2^2}}
\newcommand{\pderdertwo}[3][]{\frac{\partial^2#1}{\partial#2 \partial#3}}

\makeatletter
\def\hindawiIndent{0pc}
\def\author#1{\gdef\@author{\hskip-\dimexpr(\tabcolsep)\hskip\hindawiIndent\parbox{\dimexpr\textwidth-\hindawiIndent}{\raggedright\bfseries#1}}}

\def\title#1{\gdef\@title{\vspace*{-30pt}\raggedright\bfseries\ifx\@articleType\@empty\vspace*{0pt}\else\vspace*{0pt}\@articleType\vspace*{0pt}\\\fi#1}}
\let\@articleType\@empty \def\articletype#1{\gdef\@articleType{{\normalfont\itshape#1}}}

\makeatother

\begin{document}

\title{\large An Optical Flow-Based Approach for Minimally-Divergent Velocimetry Data Interpolation}

\author{\small\normalfont Berkay Kanberoglu\textsuperscript{1}, Dhritiman Das\textsuperscript{2}, Priya Nair\textsuperscript{3}, Pavan Turaga\textsuperscript{1,4} and
	David Frakes\textsuperscript{1,3}~\\[-3pt]\normalsize\normalfont 
	~\\\textsuperscript{1}{\small School of Electrical, Computer and Energy Engineering\unskip, Arizona State University\unskip, Tempe\unskip, 85281\unskip, USA}
	~\\\textsuperscript{2}{\small Department of Computer Science\unskip, Technical University of Munich\unskip, Munich\unskip, 80333\unskip, Germany}
	~\\\textsuperscript{3}{\small School of Biological and Health Systems Engineering\unskip, Arizona State University\unskip, Tempe\unskip, 85281\unskip, USA}
	~\\\textsuperscript{4}{\small School of Arts, Media and Engineering\unskip, Arizona State University\unskip, Tempe\unskip, 85281\unskip, USA}{\vspace*{10pt}\small ~\\Correspondence should be addressed to Berkay Kanberoglu; bkanbero@asu.edu}~\\[12pt]{\small Emails: dhritiman.das@tum.de (Dhritiman Das), pnair3@asu.edu (Priya Nair), pturaga@asu.edu (Pavan Turaga), dfrakes@asu.edu (David Frakes)}}

\date{}
\maketitle

\begin{abstract}
Three-dimensional (3D) biomedical image sets are often acquired with in-plane pixel spacings that are far less than the out-of-plane spacings between images.
The resultant anisotropy, which can be detrimental in many applications, can be decreased using image interpolation.  
Optical flow and/or other registration-based interpolators have proven useful in such interpolation roles in the past. 
When acquired images are comprised of signals that describe the flow velocity of fluids, additional information is available to guide the interpolation process.
In this paper, we present an optical-flow based framework for image interpolation that also minimizes resultant divergence 
in the interpolated data.
\end{abstract}

\section{Introduction}

Image interpolation is a fundamental problem encountered in many fields \cite{interpGR1,interpGR2,interpGR3,interpGR4,interpGR5,interpGR6,interpGR7,interpGR8,interpGR9}. There are countless scenarios wherein images are acquired at resolutions that are suboptimal for the needs of specific applications.  For example, biomedical images spanning a three-dimensional (3D) volume are often acquired with in-plane pixel spacings far less than the out-of-plane spacings between images.  This can be the case with clinical images (e.g., from computed tomography (CT) and/or magnetic resonance (MR) imaging) as well as in vitro images acquired with modalities such as particle image velocimetry (PIV) \cite{mriRef1,mriRef2,mriRef3,mriRef4,mriRef5,mriRef6,mriRef7,mriRef8}. In cases where motion estimation and registration are parts of an interpolation framework, hardware based approaches can offer solutions as well \cite{HW_ME_1,HW_ME_2,HW_ME_3,HW_ME_4,HW_ME_5,HW_ME_6,HW_ME_7}.

However, when acquired images are comprised of signals that describe the flow velocity of fluids, additional information is available to guide the interpolation process.  Specifically, the flows of an incompressible fluid into and out of an interrogation volume must be equal according to conservation of mass \cite{continuityRef}.  Quantifying the deviation from zero net flow that is entering (or alternatively leaving) an interrogation volume (i.e., divergence) thus provides a means to direct interpolation in such a way as to reconstruct more physically accurate data.  
 
Optical flow and/or other registration-based interpolators have proven useful in interpolating velocimetry data in the past \cite{regBasedinterpRef1,regBasedinterpRef2,regBasedinterpRef3,regBasedinterpRef4,regBasedinterpRef5,regBasedinterpRef6,regBasedinterpRef7,regBasedinterpRef8,regBasedinterpRef9,regBasedinterpRef10,regBasedinterpRef11,fluidSurvey}. Particle Image Velocimetry (PIV) is a technique that measures a velocity field in a fluid volume with the help of tracer particles in the fluid and specialized cameras \cite{pivBook1,pivBook2}. The default technique to determine the velocity field from the raw PIV data is a correlation analysis between two frames that were acquired by the cameras \cite{Scarano02}. This technique can be extended to 3D as well. Optical flow-based approaches have been widely used in computer vision \cite{compVis1, compVis2, compVis3, compVis4}, and they have been appealing to researchers because of the flexibility of variational approaches. Regularizers can be used for different constraints in the energy functional to be minimized. In the conventional optical flow method there are two constraints, brightness and smoothness \cite{Horn_80}. Optical flow-based methods have been promising in the area of fluid flow estimation in PIV \cite{Alvarez07,Alvarez09,Ruhnau07,Ruhnau07_2,Herlin12,Zhong17}. For example, in \cite{Alvarez09}, incompressibility of the flow is added as a constraint in the optical flow minimization problem. In \cite{Ruhnau07}, the vorticity transport equation, which describes the evolution of the fluid's vorticity over time, is used in physically consistent spatio-temporal regularization to estimate fluid motion.

Divergence and curl (vorticity) have been used in estimating optical flow previously \cite{Suter94, Gupta96, Gupta963D, Corpetti06}. In \cite{Suter94}, the smoothness constraint is decomposed into two parts, divergence and vorticity, in this way, the smoothness properties of the optical flow can be tuned. In \cite{Song91}, both incompressibility and divergence-free constraints are used in the ill-posed minimization problem to calculate a 3D velocity field from 3D Cine CT images. In \cite{Gupta963D}, a second order div-curl spline smoothness condition is employed in order to compute a 3D motion field. In \cite{Corpetti06}, a data term based on the continuity equation of fluid mechanics \cite{continuityRef} and a second order div-curl regularizer are employed to calculate fluid flow.

Here we present an optical-flow based framework for image interpolation that also minimizes resultant divergence in the interpolated data. That is, the divergence constraint attempts to minimize divergence in interpolated velocimetry data, not the divergence of the optical flow field. To our knowledge, using divergence in this way as a constraint in an optical-flow framework for image interpolation has not been investigated prior to the preliminary work presented in \cite{Kanber17}. The method is applied to PIV, computational fluid dynamics (CFD), and analytical data and results indicate that the trade-off between minimizing errors in velocity magnitude values and errors in divergence can be managed such that both are decreased below levels observed for standard truncated sinc function-based interpolators, as well as pure optical flow-based interpolators.  The proposed method thus has potential to provide an improved basis for interpolating velocimetry data in applications where isotropic flow velocity volumes are desirable, but out-of-plane data (i.e., data in different images spanning a 3D volume) can not be resolved as highly as in-plane data. 

The remainder of this paper is structured as follows. In section \ref{sec:methods}, a definition of the term optical flow will be given and a canonical optical flow method will be briefly described. This will provide a basis for the following sections as most of the work described in this paper has been built on the described method. In section \ref{sec:methods}, an optical flow-based framework for interpolating minimally divergent velocimetry data is described. The new method uses flow velocity data to guide the interpolation toward lesser divergence in the interpolated data. In section \ref{sec:results}, performance of the proposed technique is presented with experiments and simulations on real and analytical data. The results and performance of the proposed method are discussed and concluded in section \ref{sec:conclusion}.

\section{Methods}
\label{sec:methods}

\subsection{Optical Flow}

Optical flow is the apparent motion of objects in image sequences that results from relative motion between the objects and the imaging perspective. In one canonical optical flow paper \cite{Horn_80}, two kinds of constraints are introduced in order to estimate the optical flow: the {\em smoothness constraint} and the {\em brightness constancy constraint}. In this section, we give a brief overview of the original optical flow algorithm (Horn-Schunck method) and the modified algorithm that was used in this project. 

Optical flow methods estimate the motion between two consecutive image frames that were acquired at times $t$ and $t+\delta t$ . A flow vector for every pixel is calculated. The vectors represent approximations of image motion that are based in large part on local spatial derivatives. Since the flow velocity has two components, two constraints are needed to solve for it. 

\subsubsection{The Brightness Constancy Constraint}
The brightness constancy constraint assumes that the brightness of a small area in the image remains constant as the area moves from image to image. Image brightness at the point ($x,y$) in the image at time $t$ is denoted here as $I(x,y,t)$. If the point moves by $\delta x$ and $\delta y$ in time $\delta t$, then according to the brightness constancy constraint:

\begin{equation} \label{eq:brightnessconst}
\frac{dI}{dt} = 0.
\end{equation}
This can also be stated as:
\begin{equation} \label{eq:brightnessconst_E}
I(x+\delta x, y+\delta y, t+\delta t)=I(x,y,t).
\end{equation}
If we expand the left side of Eq. \ref{eq:brightnessconst_E} with a Taylor series expansion, then:
\begin{equation} \label{eq:brightnessconst_T}
I(x,y,t) + \frac{\partial I}{\partial x} \delta x + \frac{\partial I}{\partial y} \delta y + 
\frac{\partial I}{\partial t} \delta t + \dots =I(x,y,t),
\end{equation}
where the ellipsis (\dots) denotes higher order terms in the expansion. After canceling $I(x,y,t)$ from both sides of the equation:
\begin{equation} \label{eq:brightnessconst_two}
\frac{\partial I}{\partial x} \delta x +\frac{\partial I}{\partial y} \delta y +\frac{\partial I}{\partial t} \delta t + \dots = 0.
\end{equation}
We can divide this equation by $\delta t$, which leads to:
\begin{equation} \label{eq:brightnessconst_twotwo}
\frac{\partial I}{\partial x}\frac{dx}{dt}+\frac{\partial I}{\partial y}\frac{dy}{dt}
+\frac{\partial I}{\partial t} = 0.
\end{equation}
Substituting:
\begin{displaymath}
\alpha=\frac{dx}{dt} \ \ and \ \ \beta=\frac{dy}{dt},
\end{displaymath}
the brightness constraint can be written in a more compact form:
\begin{equation} \label{eq:brightnessconst_three}
I_{x}\alpha+I_{y}\beta+I_{t} = 0,
\end{equation}
where $I_{x} =\frac{\partial I}{\partial x}$, $I_{y} =\frac{\partial I}{\partial y}$, and $I_{t} =\frac{\partial I}{\partial t}$. In this form $\alpha$ and $\beta$ represent the image velocity components and ($I_{x},I_{y}$) represents the brightness gradients. 

\subsubsection{The Smoothness Constraint}
Fortunately, points from an object that is imaged in temporally adjacent frames usually have similar velocities, which results in a smooth velocity field. Leveraging this property, we can express a reasonable smoothness constraint by minimizing the sums of squares of the Laplacians of the velocity components $\alpha$ and $\beta$. The Laplacians are:

\begin{subequations}
\begin{align}
\nabla^{2} \alpha = \frac{\partial^{2} \alpha}{\partial x^{2}} + \frac{\partial^{2} \alpha}{\partial y^{2}}, \label{Lap_one}\\
\nabla^{2} \beta = \frac{\partial^{2} \beta}{\partial x^{2}} + \frac{\partial^{2} \beta}{\partial y^{2}}. \label{Lap_two}
\end{align}
\end{subequations}

\subsubsection{Minimization}
Optical flow assumes constant brightness and smooth velocity over the whole image. The two constraints described above are used to formulate an energy functional to be minimized:

\begin{equation} 
\epsilon = \iint\left[\left(I_{x}\alpha+I_{y}\beta+I_{t}\right)^{2} + \lambda^{2}\left(\frac{\partial^{2} \alpha}{\partial x^{2}} + \frac{\partial^{2} \alpha}{\partial y^{2}} +  \frac{\partial^{2} \beta}{\partial x^{2}} + \frac{\partial^{2} \beta}{\partial y^{2}}\right)\right]dx\, dy.\label{eq:energyf}
\end{equation}
Using variational calculus, the Euler-Lagrange equations can be determined for this problem. Those equations need to be solved for each pixel in the image. Iterative methods are suitable to solve the equations since it can be very costly to solve them simultaneously. The iterative equations that minimize Eq. \ref{eq:energyf} are:

\begin{subequations}
\begin{align}
\alpha^{n+1} = \bar{\alpha}^{n} - \frac{I_{x}\left[I_{x} \bar{\alpha}^{n} + I_{y} \bar{\beta}^{n} + I_{t}\right]}{\lambda^{2} + I_x^2 + I_y^2},
\label{it_u} \\
\beta^{n+1} = \bar{\beta}^{n} - \frac{I_{y}\left[I_{x} \bar{\alpha}^{n} + I_{y} \bar{\beta}^{n} + I_{t}\right]}{\lambda^{2} + I_x^2 + I_y^2},
\label{it_v}
\end{align}
\end{subequations}
where \emph{n} denotes the iteration number and $\bar{\alpha}^{n}$ and $\bar{\beta}^{n}$ denote neighborhood averages of $\alpha^{n}$ and $\beta^{n}$. More detailed information on the method can be found in \cite{Horn_80}.

\subsection{Optical Flow with Divergence Constraint}
\label{sec:divergenceOF}

\subsubsection{Continuity Equation}
\begin{figure}[h]
	\centering
	\includegraphics[width=3in]{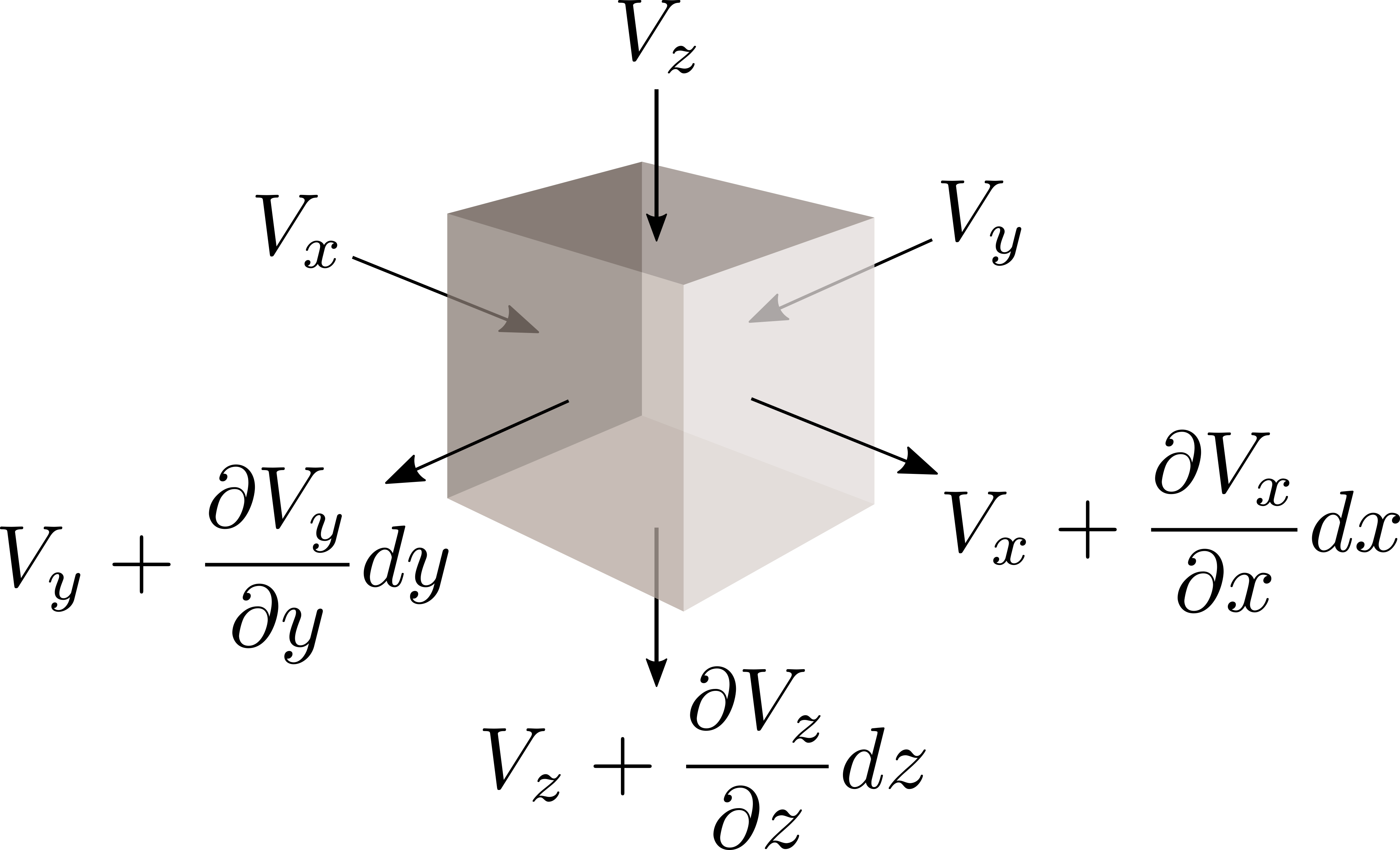}
	\caption{Change in flow velocity of a sample voxel. \label{fig:continuityeq}}
\end{figure}

According to the continuity equation in fluid dynamics, the rate of mass entering a system is equal to the rate of the mass leaving the system \cite{continuityRef}. The differential form of the equation is:

\beq \label{CEdif}
\pder[\rho]{t} + \nabla \cdot (\rho \overrightarrow{\mathbf{u}}) = 0,
\eeq
where $\rho$ is the fluid density, $t$ is time and $\overrightarrow{\mathbf{u}}$ is the velocity vector field. In the case of incompressible flow, $\rho$ becomes constant and the continuity equation takes the form: 

\beq \label{CE}
\nabla \cdot \overrightarrow{\mathbf{u}} = \pder[V_x]{x} + \pder[V_y]{y} + \pder[V_z]{z} = 0. 
\eeq
This means that the divergence of the velocity field is zero in the case of incompressible flow. Figure \ref{fig:continuityeq} shows the change in flow velocity of a voxel.
 
\subsubsection{Symmetric Setup}
\begin{figure}[t]
	\centering
	\includegraphics[width=4.5in]{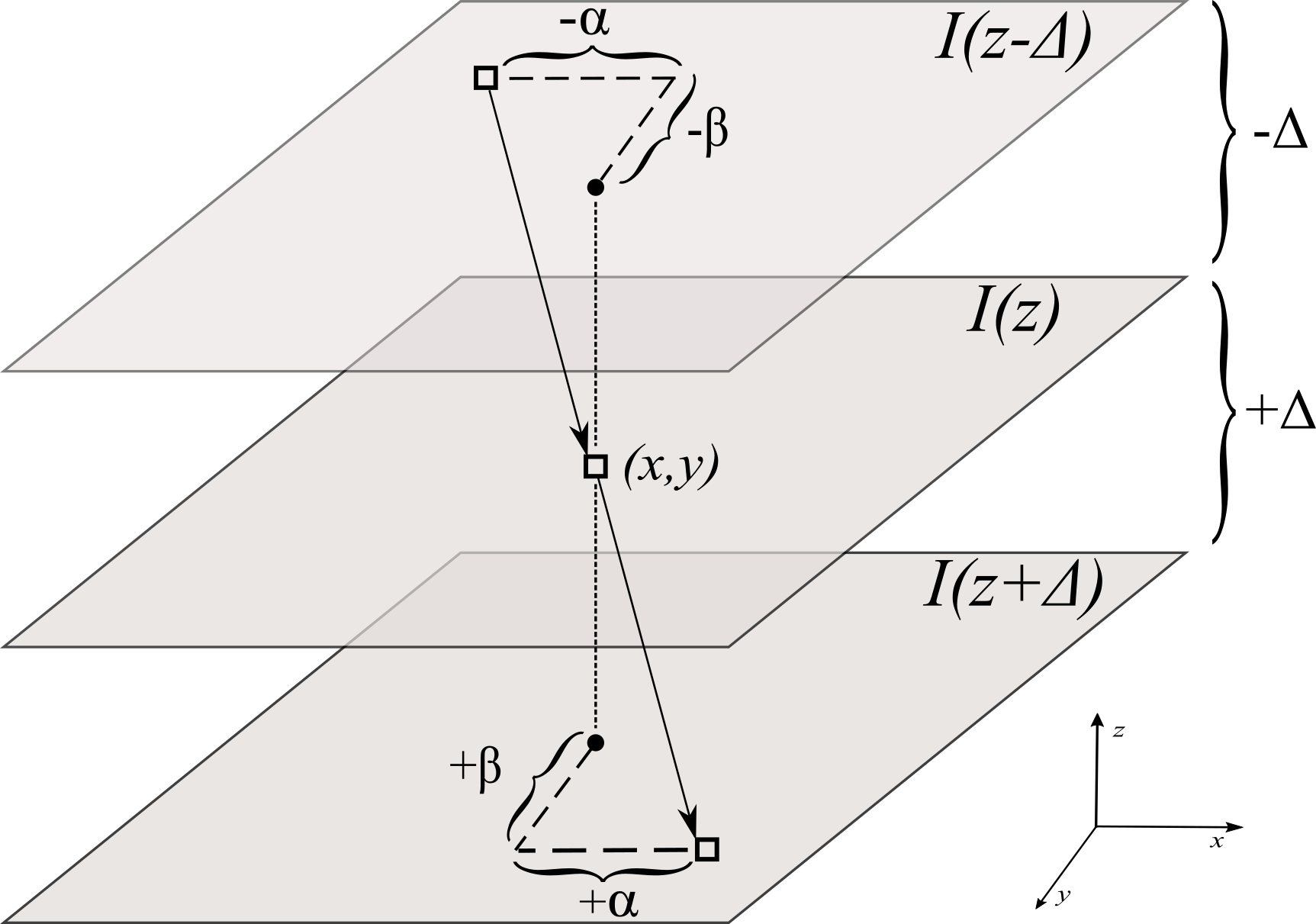}
	\caption{Illustration of the symmetric interpolation setup. \label{fig:stackedSlices}}
\end{figure}

For the new method, a symmetric interpolation setup is proposed as shown in Figure \ref{fig:stackedSlices}. In the figure, upper and lower slices are from the dataset and the interpolated slice is in the middle. 

\beq \label{eq:brightnessconst_I}
I(x+\alpha, y+\beta, z+\Delta)=I(x-\alpha, y-\beta, z-\Delta).
\eeq
In this section, $I(x,y,t)$ denotes the velocity magnitude image and $\overrightarrow{\mathbf{V}}$ denotes the velocity vector components (i.e., $V_x$,$V_y$,$V_z$). If one approximates the expressions with Taylor expansion around the points $(x,y)$, we get:
\begin{subequations} 
	\begin{align}
	I(x+\alpha, y+\beta, z+\Delta) = I(x,y,z+\Delta) + \frac{\partial I(x,y,z+\Delta)}{\partial x} \alpha + \frac{\partial I(x,y,z+\Delta)}{\partial y} \beta + ... \label{tayexp+} \; ,\\
	I(x-\alpha, y-\beta, z-\Delta) = I(x,y,z-\Delta) - \frac{\partial I(x,y,z-\Delta)}{\partial x} \alpha - \frac{\partial I(x,y,z-\Delta)}{\partial y} \beta + ... \label{tayexp-} \; .
	\end{align}
\end{subequations}
After substituting Eqs. \ref{tayexp+} and \ref{tayexp-} into Eq. \ref{eq:brightnessconst_I}, terms can be arranged to obtain the new brightness constraint:
\beq \label{newbrightnessconst}
\begin{split}
& \left[ I(x, y, z+\Delta)-I(x, y, z-\Delta) \right] 
\\
&\quad	+ \alpha \left[ \frac{\partial I(x,y,z+\Delta)}{\partial x} + \frac{\partial I(x,y,z-\Delta)}{\partial x} \right] 
\\
&\qquad	+ \beta \left[  \frac{\partial I(x,y,z+\Delta)}{\partial y} + \frac{\partial I(x,y,z-\Delta)}{\partial y} \right] = 0.
\end{split}
\eeq

In the next step, we aim to minimize the divergence of the interpolated slice. Ideally, the divergence equation of the interpolated slice should be used:
\beq \label{divequation_interpslice}
\nabla \cdot \overrightarrow{\mathbf{V}}(z) = \pder[V_x(x, y, z)]{x} + \pder[V_y(x, y, z)]{y} + \pder[V_z(x, y, z)]{z} = 0.
\eeq
Since this information is unavailable, to generate the middle slice with as little divergence as possible, we can use the fact that:
\beq \label{divequality}
\nabla \cdot \overrightarrow{\mathbf{V}}(z) = \nabla \cdot \overrightarrow{\mathbf{V}}(z+\Delta) = \nabla \cdot \overrightarrow{\mathbf{V}}(z-\Delta) = 0.
\eeq
which leads to the following constraint by using the divergence expressions of the two outer slices, $I(z-\Delta)$ and $I(z+\Delta)$:

\beq \label{divconst2}
\begin{split}
& \pder[V_x(x+\alpha, y+\beta, z+\Delta)]{x} + \pder[V_y(x+\alpha, y+\beta, z+\Delta)]{y} + \pder[V_z(x+\alpha, y+\beta, z+\Delta)]{z}
\\
&\qquad + \pder[V_x(x-\alpha, y-\beta, z-\Delta)]{x} + \pder[V_y(x-\alpha, y-\beta, z-\Delta)]{y} + \pder[V_z(x-\alpha, y-\beta, z-\Delta)]{z} = 0.
\end{split}
\eeq
Using Taylor expansion on Eq. \ref{divconst2} yields:
\beq \label{divconst2Taylor1}
\begin{split} 	
& \left[ \pder[V_x(z+\Delta)]{x}  + \pder[V_x(z-\Delta)]{x} + \pder[V_y(z+\Delta)]{y}  + \pder[V_y(z-\Delta)]{y} + \pder[V_z(z+\Delta)]{z}  + \pder[V_z(z-\Delta)]{z} \right]
\\
& + \alpha \left[ \pderder[V_x(z+\Delta)]{x} - \pderder[V_x(z-\Delta)]{x} + \pderdertwo[V_y(z+\Delta)]{x}{y} - \pderdertwo[V_y(z-\Delta)]{x}{y} + \pderdertwo[V_z(z+\Delta)]{x}{z} - \pderdertwo[V_z(z-\Delta)]{x}{z} \right]
\\
& + \beta \left[ \pderdertwo[V_x(z+\Delta)]{y}{x} - \pderdertwo[V_x(z-\Delta)]{y}{x} + \pderder[V_y(z+\Delta)]{y} - \pderder[V_y(z-\Delta)]{y} + \pderdertwo[V_z(z+\Delta)]{y}{z} - \pderdertwo[V_z(z-\Delta)]{y}{z} \right] = 0.
\end{split}
\eeq
In Eq. \ref{divconst2Taylor1}, we need the derivatives of $V_z(z+\Delta)$ and $V_z(z-\Delta)$ in the z-direction. Calculating these derivatives in the z-direction would require additional outer slices. To simplify this requirement, we can expand $V_z(x+\alpha, y+\beta, z+\Delta)$ and $V_z(x-\alpha, y-\beta, z-\Delta)$ around the points $(x,y,z)$ and obtain the following,
\begin{subequations}  
	\begingroup
	\makeatletter\def\f@size{11}\check@mathfonts
	\def\maketag@@@#1{\hbox{\m@th\normalfont\normalsize#1}} \makeatother
	\begin{align}
		\pder[V_z(x+\alpha, y+\beta, z+\Delta)]{z} = \pder[V_z(x, y, z)]{z} + \alpha \pderdertwo[V_z(x, y, z)]{x}{z} + \beta \pderdertwo[V_z(x, y, z)]{y}{z} + \Delta \pderder[V_z(x, y, z)]{z} + ... \label{Vzexpand1} \\
		\pder[V_z(x-\alpha, y-\beta, z-\Delta)]{z} = \pder[V_z(x, y, z)]{z} - \alpha \pderdertwo[V_z(x, y, z)]{x}{z} - \beta \pderdertwo[V_z(x, y, z)]{y}{z} - \Delta \pderder[V_z(x, y, z)]{z} + ... \label{Vzexpand2}
	\end{align} 
	\endgroup
\end{subequations}

Using Eqs. \ref{Vzexpand1} and \ref{Vzexpand2} in Eq. \ref{divconst2}, we obtain the new divergence constraint that doesn't require additional slices for the z-direction derivative, 
\beq \label{divconst2Taylor2}
\begin{split}
& \left[ \pder[V_x(z+\Delta)]{x}  + \pder[V_x(z-\Delta)]{x} + \pder[V_y(z+\Delta)]{y}  + \pder[V_y(z-\Delta)]{y} + 2 \pder[V_z]{z} \right]
\\
&\quad + \alpha \left[ \pderder[V_x(z+\Delta)]{x} - \pderder[V_x(z-\Delta)]{x} + \pderdertwo[V_y(z+\Delta)]{x}{y} - \pderdertwo[V_y(z-\Delta)]{x}{y} \right]
\\
&\quad + \beta \left[ \pderdertwo[V_x(z+\Delta)]{y}{x} - \pderdertwo[V_x(z-\Delta)]{y}{x} + \pderder[V_y(z+\Delta)]{y} - \pderder[V_y(z-\Delta)]{y} \right] = 0.
\end{split}
\eeq
Combining Eqs. \ref{newbrightnessconst}, \ref{divconst2Taylor2} and the optical flow smoothness constraint, we obtain the new energy functional that needs to be minimized,
\beq 
\epsilon = \iint\left[H_{x}\alpha+H_{y}\beta+H_{z}\right]^{2} + \gamma^{2}\left[ D_x\alpha + D_y\beta + D_z\right]^2 + \lambda^{2}\left[\|\nabla\alpha\|^{2} + \|\nabla\beta\|^{2}\right]dx\, dy
\eeq \label{eq:newenergyf}
where 
\begin{align*}
	H_x &= \left[ \frac{\partial I(x,y,z+\Delta)}{\partial x} + \frac{\partial I(x,y,z-\Delta)}{\partial x} \right]
	\\
	H_y &= \left[  \frac{\partial I(x,y,z+\Delta)}{\partial y} + \frac{\partial I(x,y,z-\Delta)}{\partial y} \right]
	\\
	H_z &= \left[ I(x, y, z+\Delta)-I(x, y, z-\Delta) \right]
	\\
	D_x &= \left[ \pderder[V_x(z+\Delta)]{x} - \pderder[V_x(z-\Delta)]{x} + \pderdertwo[V_y(z+\Delta)]{x}{y} - \pderdertwo[V_y(z-\Delta)]{x}{y} \right]
	\\ 
	D_y &= \left[ \pderdertwo[V_x(z+\Delta)]{y}{x} - \pderdertwo[V_x(z-\Delta)]{y}{x} + \pderder[V_y(z+\Delta)]{y} - \pderder[V_y(z-\Delta)]{y} \right]
	\\
	D_z &=\left[ \pder[V_x(z+\Delta)]{x}  + \pder[V_x(z-\Delta)]{x} + \pder[V_y(z+\Delta)]{y}  + \pder[V_y(z-\Delta)]{y} + 2 \pder[V_z]{z} \right]
\end{align*}	
Using variational calculus, the Euler-Lagrange equations can be determined for this problem. They need to be solved for each pixel in the image. The iterative equations that minimize the solutions are given by,

\begin{subequations}
	\begin{align}
	\alpha^{n+1} = \bar{\alpha}^{n} - \frac{A_1 \bar{\alpha}^{n} + B_1 \bar{\beta}^{n} + \gamma^{2}C_1 + \lambda^{2}C_2}{\gamma^{2}D_1 + \lambda^{2}D_2},
	\label{newit_alpha} \\
	\beta^{n+1} = \bar{\beta}^{n} - \frac{A_2 \bar{\alpha}^{n} + B_2 \bar{\beta}^{n} + \gamma^{2}C_3 + \lambda^{2}C_4}{\gamma^{2}D_3 + \lambda^{2}D_4},
	\label{newit_beta}
	\end{align}
\end{subequations}
where $n$ denotes the iteration number and $\bar{\alpha}^{n}$ and $\bar{\beta}^{n}$ denote neighborhood averages of $\alpha^{n}$ and $\beta^{n}$. 
The coefficient expressions in Eqs. \ref{newit_alpha} and \ref{newit_beta} are given as

\begin{align*}
A_1 &= \gamma^{2}\left(H_xD_y - H_yD_x\right)^{2} + \lambda^{2}\left(H_x^{2} + \gamma^{2}D_x^{2}\right) \\
B_1 &= \lambda^{2}\left(H_xH_y + \gamma^{2}D_xD_y\right) \\
C_1 &= H_xH_zD_y^{2} + H_y^{2}D_xD_z - H_yH_zD_xD_y - H_xH_yD_yD_z \\
C_2 &= H_xH_z + \gamma^{2}D_xD_z\\
D_1 &= \left(H_xD_y-H_yD_x\right)^{2}\\
D_2 &= \left(H_x^2 + H_y^2 +\lambda^{2} + \gamma^{2}D_x^{2} + \gamma^{2}D_y^{2}\right)\\
A_2 &= B_1\\
B_2 &= \gamma^{2}\left(H_xD_y - H_yD_x\right)^{2} + \lambda^{2}\left(H_y^{2} + \gamma^{2}D_y^{2}\right) \\
C_3 &= H_yH_zD_x^{2} + H_x^{2}D_yD_z - H_xH_zD_xD_y - H_xH_yD_xD_z\\
C_4 &= H_yH_z + \gamma^{2}D_yD_z\\
D_3 &= D_1\\
D_4 &= D_2
\end{align*}
The numerical scheme to solve the Euler-Lagrange equations is based on the solution laid out in \cite{Horn_80}. More detailed information on the steps of the derivation can be found in the appendix.

There have been several studies that attempt to improve the performance of optical flow techniques and computation schemes \cite{OF_comp1,OF_comp2,OF_comp3,OF_comp4,OF_comp5,OF_comp6,OF_comp7,OF_comp8,OF_comp9}. 
For example, in \cite{OF_comp3} non-linear convex penalty functions are used for the constraints in the optical flow energy functional. The approach uses numerical approximations to obtain a sparse linear system of equations from the highly nonlinear Euler-Lagrange equations. The resulting linear system of equations can be solved with numerical methods like Gauss-Seidel, which is similar to Jacobi method, or successive over-relaxation (SOR), which is a Gauss-Seidel variant. 
Another improvement to variational optical flow computation is presented in \cite{OF_comp4}. The approach uses a multigrid numerical optimization method and because of its speedup gains, it can be used in real-time. 
After all these advances, in \cite{OF_comp1}, it was argued that the typical formulation of optical flow has changed little and most of the advances have been mainly numerical optimization and implementation techniques, and robustness functions. This is also true for the proposed method as well. The optical flow portion of this interpolation framework is closely related to the Horn-Schunck method. The derived numerical scheme to solve the equations enhances this notion while its implementation is straightforward and simple. For example, setting the divergence coefficient $\gamma$ to $0$ in Eqs. \ref{newit_alpha} and \ref{newit_beta} reduces the solutions to Horn-Schunck solutions. The numerical scheme is also sufficient for velocimetry data because unlike in many other types of images, stark discontinuities are unexpected in velocimetry images at Reynolds numbers on the order of biomedical flows.

\newpage
\clearpage
\begin{figure}[h!]
	\centering
	\includegraphics[width=3.5in]{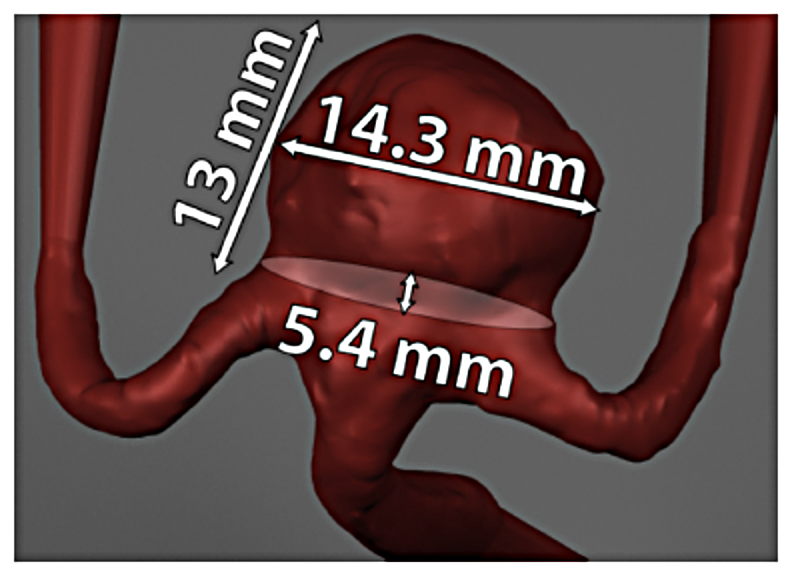}
	\caption{Dimensions of the aneurysm \label{fig:aneurysmIm}}
\end{figure}

\subsection{PIV Setup}
\begin{figure}[htb!]
	\centering
	\includegraphics[width=4in]{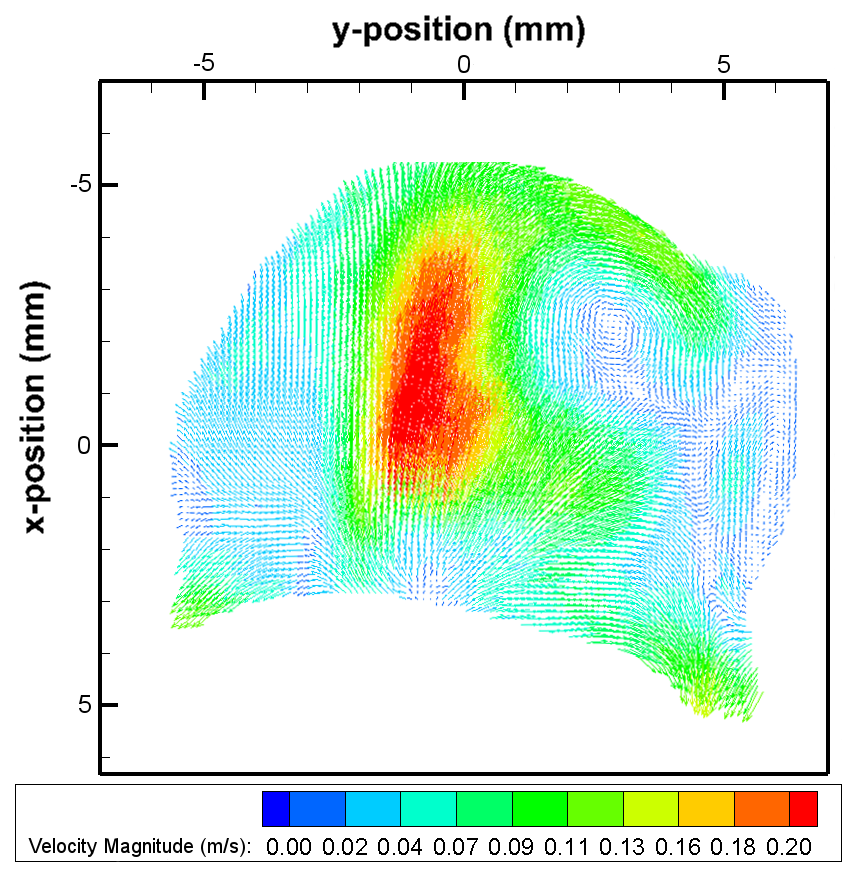}
	\caption{Example flow slice from the PIV experiments. \label{fig:aneurysmFlow}}
\end{figure}

The testing datasets were acquired using particle image velocimetry, an optical experimental flow measurement technique. PIV data acquisition and processing generally consists of the following steps: (1) computational modeling, (2) physical model construction, (3) particle image acquisition, (4) PIV processing, and (5) data analysis. The testing datasets were acquired for an in-vitro model of a cerebral aneurysm. Patient-specific computed tomography (CT) images were first segmented and reconstructed to obtain the computational cerebral aneurysm model as shown in Figure~\ref{fig:aneurysmIm}. The computational model was then translated into an optically clear, rigid urethane model using a lost-core manufacturing methodology. The physical model was connected to a flow loop consisting of a blood analog solution seeded with 8 $\mu$m fluorescent microspheres. Fluid flow through the physical model was controlled at specific flow rates (3, 4 and 5 mL/s). PIV was performed using a FlowMaster 3D Stereo PIV system (LaVision, Ypsilanti, MI), where the fluorescent particles were illuminated with a 532 nm dual-pulsed Nd:YAG laser at a controlled rate, while two CCD cameras captured the images across seven parallel planes (or slices) within the aneurysmal volume. A distance of 1 mm separated the planes. Two hundred image pairs, at each flow rate and slice, were acquired at 5 Hz. The image pairs were processed using a recursive cross-correlation algorithm using Davis software (LaVision, Ypsilanti, MI) to calculate the velocity vectors within region of interest (i.e., the aneurysm). Initial and final interrogation window sizes of 32 by 32 pixels and 16 by 16 pixels, respectively, were used. Detailed explanation of the experimental process can be found in \cite{aneurysmRef}. A sample experimental model is shown in Figure~\ref{fig:aneurysmFlow}. 

The proposed algorithm was developed in MATLAB (Mathworks, Inc). Since the proposed algorithm has two separate terms for divergence and smoothness, different combinations of coefficients can be used for the terms. However, in order to get a clear idea about the performance of the method only one set of parameters were used in the simulations. The divergence term’s coefficient $\gamma$ was set to 150. From previous tests, it was seen that the proposed method performed better when a relatively large $\gamma$ was used while keeping the smoothness coefficient $\lambda$ small. The smoothness coefficient $\lambda$ was set to 1. The same smoothness coefficient was also used for the Horn-Schunck based method. The iterations for both methods were set to 2000. Each PIV dataset used in testing had 7 slices. The slices were originally 154x121. They were cropped and zero-padded to reach 128x128. The size of the region where MSE and divergence were calculated is 110x110. Even though there are 7 slices in each dataset, only 3 slices were reconstructed from the datasets. These are slices 3, 4 and 5. Two different spacing steps were used between the slices. The first one is $\Delta$z=2 where the neighboring slices z-1 and z+1 were used to reconstruct the middle slice. The second one is $\Delta$z=4 where slices z-2 and z+2 were used for the interpolation, e.g., slices 1 and 5 were used to reconstruct slice 3. The method was tested against linear interpolation and an implementation of Horn-Schunck optical flow based interpolation.

\subsection{Analytical Datasets}
The method was tested with a 3D divergence-free analytical dataset and a CFD data set with turbulent flow. The analytical dataset is given below.
\begin{subequations}
	\begin{align}
	V_x &= 0.3 y^{2} + 0.15 x^{2} \\
	V_y &= 0.3 \left(1-x^{2}\right) \left(y-1\right) - 0.3 y x \\
	V_z &= -0.3 \left(1-x^{2}\right)z
	\end{align}
\end{subequations}
Out-of-plane distance was kept much higher than the in-plane resolution. 
In order to assess the robustness of the proposed method, each velocity field was perturbed by Gaussian noise. The noise had zero mean and standard deviation of 10\% of the maximum velocity in each velocity field. 

\subsection{Computational Fluid Dynamics (CFD) Simulations}
The original computational aneurysm model was imported into ANSYS ICEM (ANSYS, Canonsburg, PA), where the inlet and outlets of the aneurysm model were extruded. 
After meshing was performed to discretize blood volumes into tetrahedrons, the final mesh was imported into ANSYS Fluent where the blood volume was modeled as an 
incompressible fluid with the same density and viscosity as the blood analog solution used in experiments. 
The vessel wall was assumed to be rigid, and a no-slip boundary condition was applied at the walls. A steady flat 4ml/s flow profile was applied at the inlet of the model, 
and zero pressure boundary conditions were imposed at the outlets. The overall CFD approach has been described previously in \cite{aneurysmRef, cfdRef}.


\section{Results} 
\label{sec:results} 

\begin{figure}[ht!]
	\centering
	\includegraphics[width=6.5in]{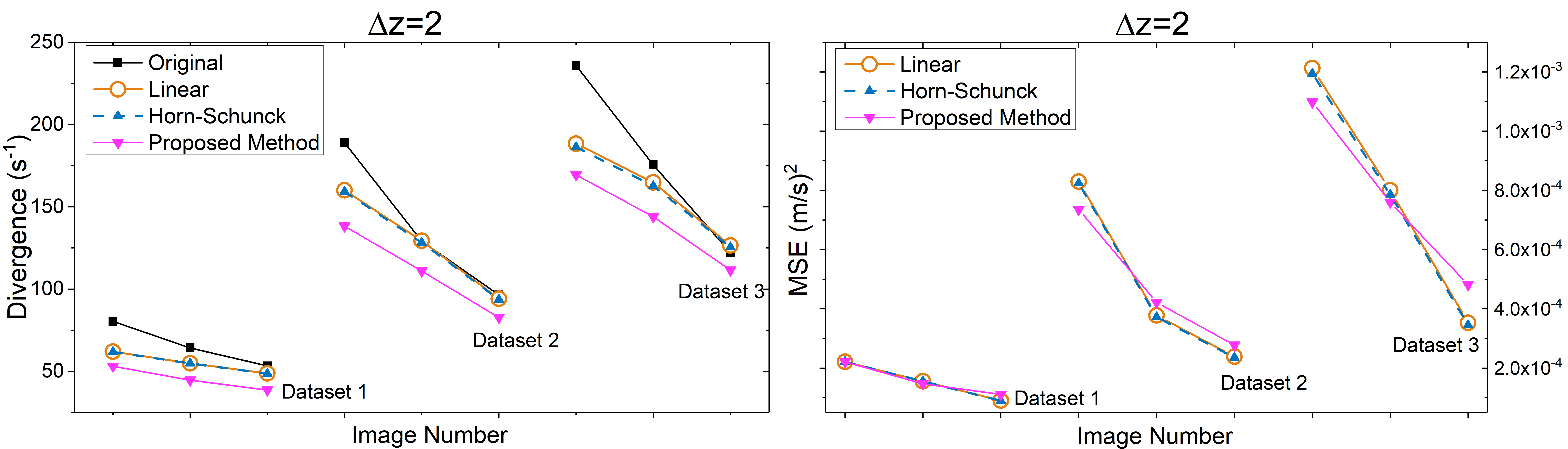}
	\caption{Divergence and MSE comparisons when slice distance is 2mm. \label{fig:divmseCombz2}}
\end{figure}

Figure~\ref{fig:divmseCombz2} shows divergence and MSE comparison graphs when $\Delta$z=2. The proposed method consistently achieves lower divergence values than the Horn-Schunck-based interpolation whereas the MSE values vary between better and worse values. On average, divergence values were 11\% lower than the Horn-Schunck-based interpolation. In some cases, the proposed method achieves up to 20\% lower divergence values.

\begin{figure}[t!]
	\centering
	\includegraphics[width=6.5in]{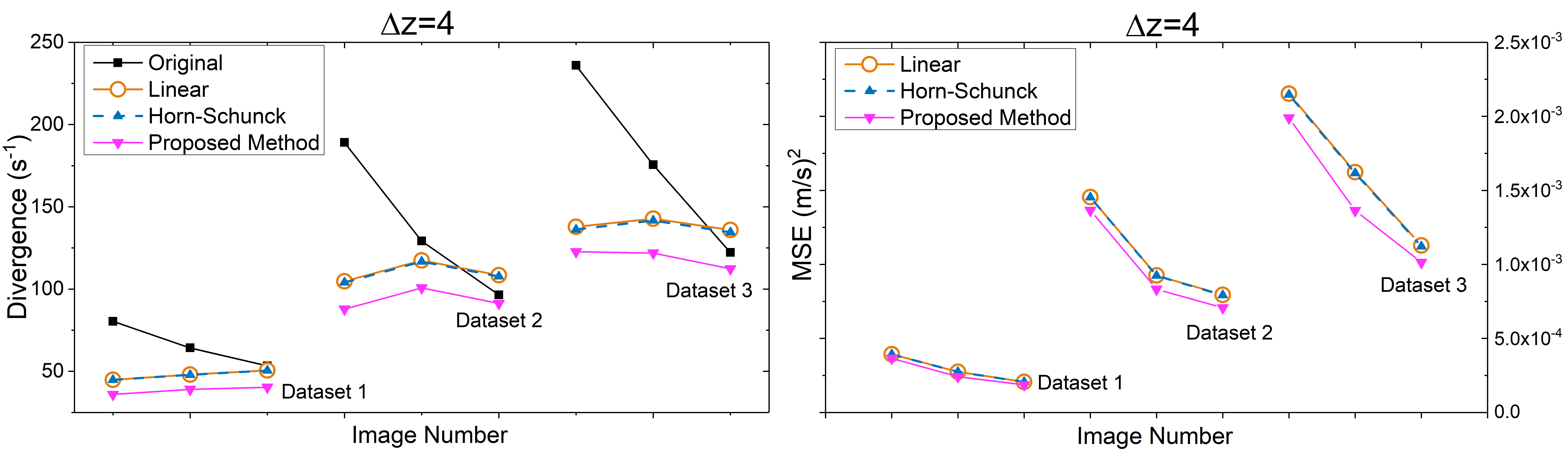}
	\caption{Divergence and MSE comparisons when slice distance is 4mm. \label{fig:divmseCombz4}}
\end{figure}

\begin{figure}[t!]
	\centering
	\includegraphics[width=6.5in]{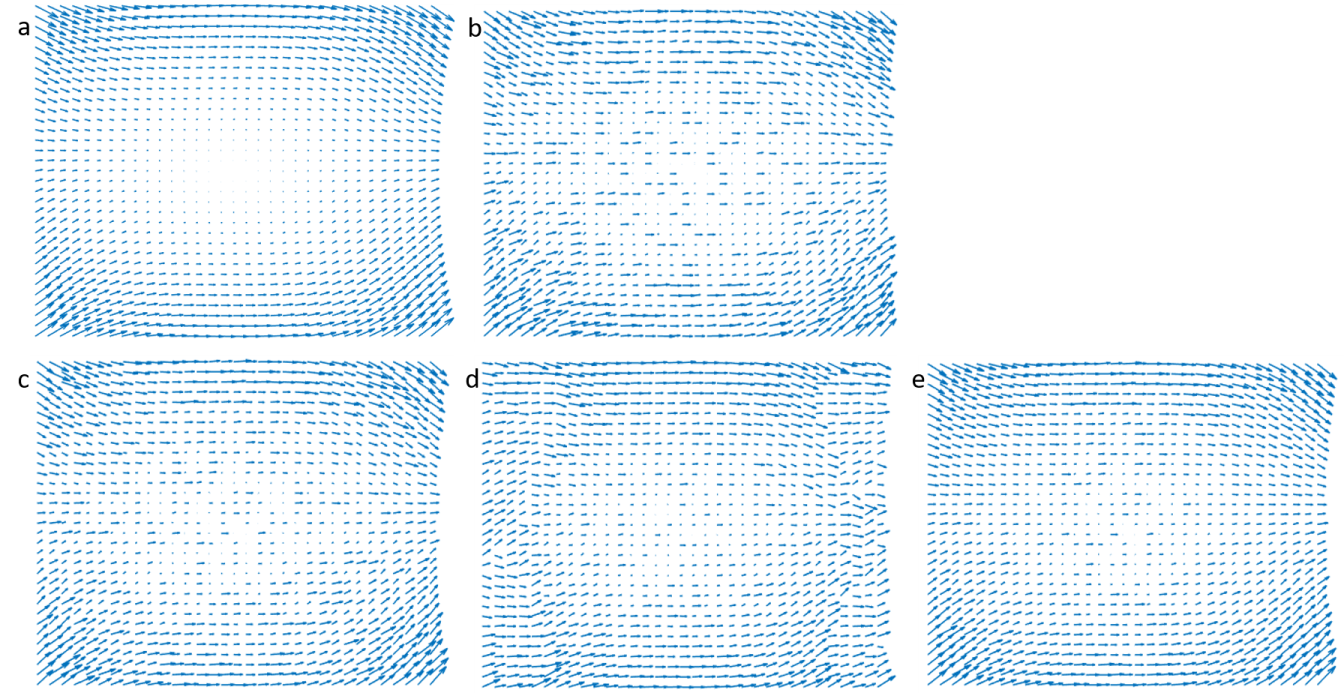}
	\caption{Plotted $V_x$ and $V_y$ components of the 3D analytical divergence-free vector field. a) Original, b) Gaussian noise added, c) Linear interpolation, d) Horn-Schunck based interpolation, e) Proposed method. Note that the proposed method is able to achieve a smoother velocity field in the corners of the interpolated data. \label{fig:analyticalComp}}
\end{figure}

\begin{figure}[t!]
	\centering
	\includegraphics[width=6.5in]{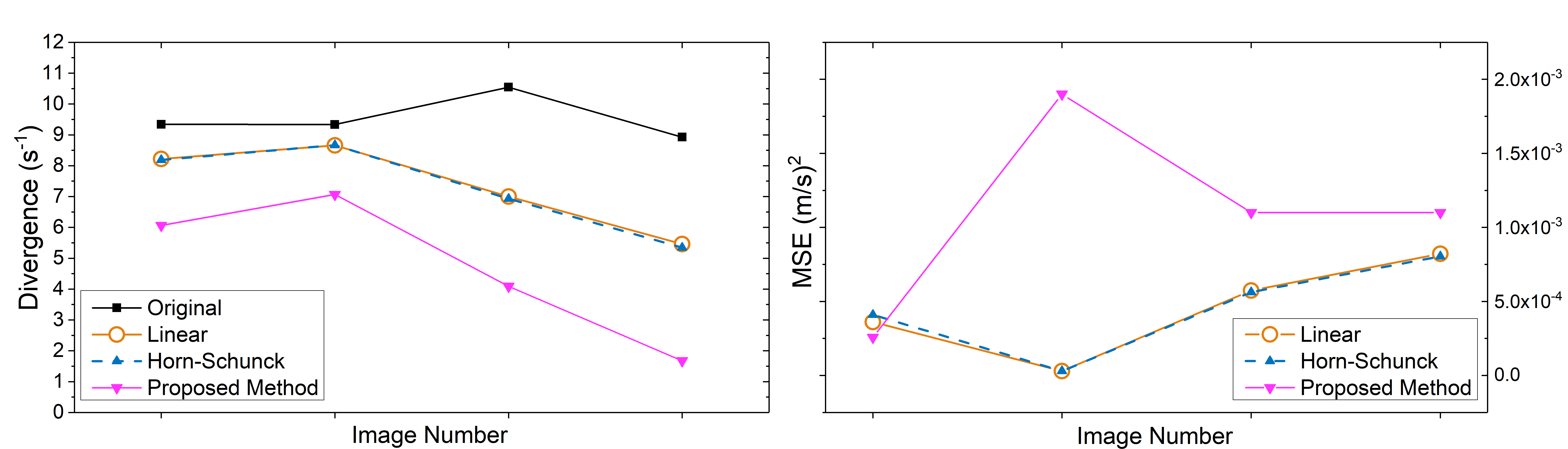}
	\caption{Divergence and MSE comparisons for the CFD dataset. \label{fig:divmseCombCFD}}
\end{figure}

Figure~\ref{fig:divmseCombz4} shows divergence and MSE comparison graphs when $\Delta$z=4. In this case, the proposed method consistently achieves lower divergence and MSE values than the other tested methods. 
Figure~\ref{fig:analyticalComp} shows original, noisy, and interpolated slices from the analytical dataset for comparison. In the figure, only $V_x$ and $V_y$ components were plotted to show the effect of the divergence term. 
In Fig.~\ref{fig:divmseCombCFD}, it can be seen that the proposed algorithm reduces divergence while the MSE is increased in the CFD dataset. 

\begin{figure}[t]
	\centering
	\includegraphics[width=6.5in]{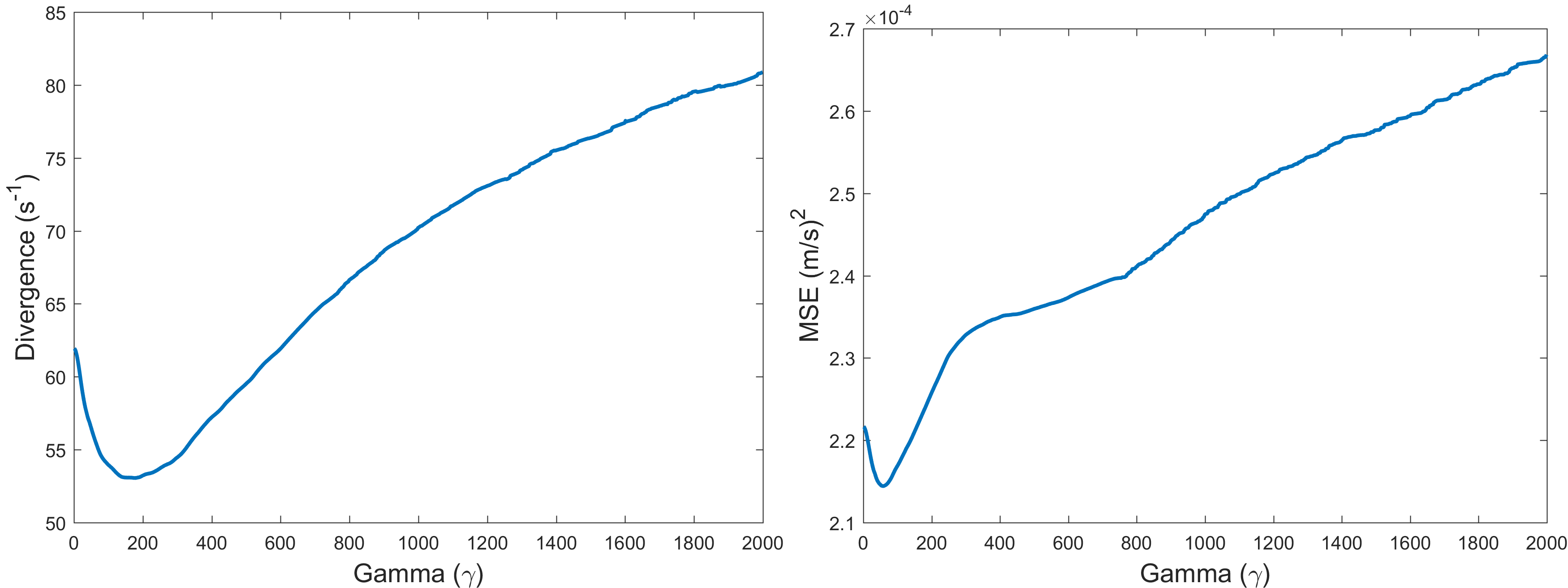}
	\caption{Divergence and MSE profiles of the proposed method as $\gamma$ is increased linearly while $\lambda=1$.\label{fig:sweepCombGamma}}
\end{figure}

\begin{figure}[t]
	\centering
	\includegraphics[width=6.5in]{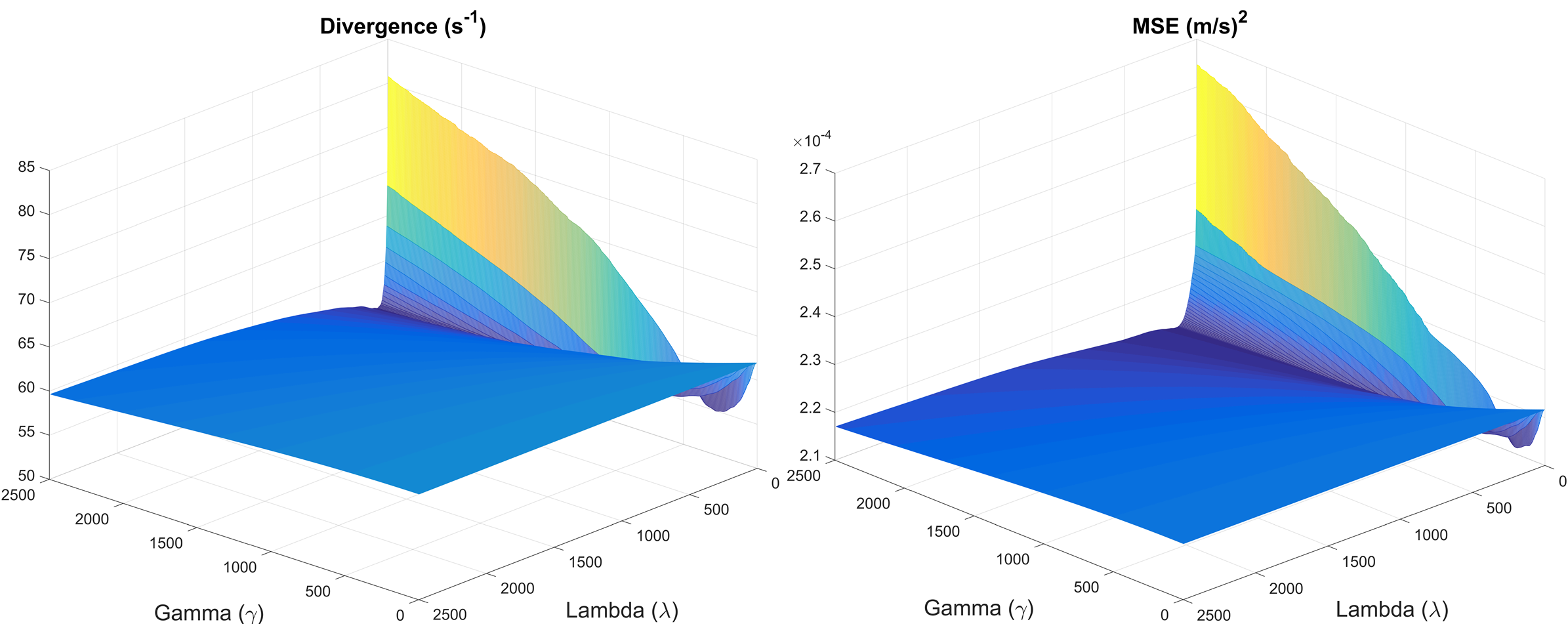}
	\caption{Divergence and MSE profiles of the proposed method as $\gamma$ and $\lambda$ are increased linearly from 0.1 to 2500.\label{fig:sweepComb2}}
\end{figure}

The graphs in Figure~\ref{fig:sweepCombGamma} show the behavior of the proposed method as the divergence coefficient $\gamma$ increases linearly. In this simulation, the smoothness coefficient $\lambda$ was kept constant ($\lambda$=1). The graphs are taken from the PIV dataset. The divergence graph profiles were consistent across different images and three datasets. The MSE graph profiles may differ slightly from the divergence graph profiles across different datasets, but MSE always increased with increasing $\gamma$. The coefficient values tested were from 0 to 2000. The profiles shown in the figure show that there needs to be a balance between the divergence and smoothing terms. The graphs in the figure are consistent with profiles of other published $\ell_{2}$-based regularization methods \cite{sweepRef1,sweepRef2}. 
Figure~\ref{fig:sweepComb2} shows the behavior of the proposed method as $\gamma$ and $\lambda$ increase linearly.

The computational cost of obtaining flow vectors with the proposed method is similar to that of the Horn-Schunck approach. Even though the iterative solutions of the proposed method employ several terms, these need to be computed only once and can be reduced to a simpler form that is similar to the Horn-Schunck solutions. Both approaches took around 0.1 seconds to obtain an optical flow field on a single core of an Intel dual core CPU (i7-6500U @ 2.50GHz). 

Another parameter that could affect the divergence, MSE, and computational cost was the iteration number. We ran simulations with different iteration numbers and noted that the divergence and MSE results seem to stabilize after 200 iterations. Higher iteration number mostly had an effect on the computation time. 

\newpage
\clearpage
\section{Discussion and Conclusions}
\label{sec:conclusion} 

A new optical flow-based framework for image interpolation that also reduces divergence is proposed. The new method
uses flow velocity data to guide the interpolation toward lesser divergence in the interpolated data. In addition to the symmetric
interpolation setup, the method introduces a new divergence term into the canonical optical flow method. The method is 
applied to PIV, analytical, and CFD data. The method was tested against linear interpolation and the Horn-Schunck optical flow method 
since it uses a similar formulation as the Horn-Schunck method. The proposed method applies a symmetric interpolation setup and 
considers a new divergence term in addition to the brightness and smoothness terms in the energy functional. 

In order to test the effects of the divergence term, both the Horn-Schunck and proposed methods were subject to the same smoothness coefficient. When tested on
the noisy analytical data, the proposed method achieved a smoother and less noisy interpolated velocity field. 

The proposed method was also applied to the PIV data with different values of smoothness and divergence term coefficients, $\alpha$ and $\gamma$, respectively. 
Results indicate that the tradeoff between minimizing errors in velocity magnitude values 
and errors in divergence can be managed such that both are decreased below levels observed for standard truncated sinc 
function-based interpolators as well as pure optical flow-based interpolators. The divergence term coefficient, $\gamma$, needs to be large enough to reduce divergence in the interpolated data but not so large as to dominate the energy functional and introduce errors into the final interpolated velocity field. The effect of the iteration number on the divergence and MSE numbers was found to be minimal after 200 iterations. The computational cost of the method was similar to that of the Horn-Schunck based approach.

The method uses a numerical scheme that is well-known and straightforward. It's true that a more recent optical flow computation scheme may lead to performance gains in quality and/or speed-up. Methods presented in \cite{OF_comp3} and \cite{OF_comp4} have become popular because of their speed, simplicity, and flexibility. Adoptation of recent numerical optimization and implementation techniques will be explored for future research.  

The proposed method has potential to improve the interpolation of velocimetry data when it's difficult achieve an out-of-plane resolution close to the in-plane resolution. The results also indicate that the effect of the new divergence term in the optical flow functional can be appreciated better as the distance between the 
interpolated slice and the neighboring slices increases. It was noted that the proposed method outperforms the tested methods in both divergence and MSE values 
when the slice distance was increased. When the slice distance is small, the proposed method achieves lower divergence than the other methods while achieving similar
MSE values.  

\titlespacing*{\section}{0pt}{10pt}{8pt}
\section*{Data Availability}
The PIV data used to support the findings of this study are available from the corresponding author upon request. Sample code can be found at \url{https://github.com/berk-github/OF_interp}. 
\section*{Conflicts of Interest}
The authors declare that there are no conflicts of interest regarding the publication of this paper.
\section*{Funding Statement}
This work was supported by the National Science Foundation [1512553].
\section*{Acknowledgments}
A portion of this work was presented as an abstract at the 2017 IEEE 14\textsuperscript{th} International Symposium on Biomedical Imaging (ISBI 2017).

\newpage
\appendix
\titlespacing*{\section} {0pt}{3.5ex plus 1ex minus .2ex}{2.3ex plus .2ex}
\section{Appendix} 
\label{appendix}
\beq 
\epsilon = \iint\left[I_{x}\alpha+I_{y}\beta+I_{z}\right]^{2} + \gamma^{2}\left[ D_x\alpha + D_y\beta + D_z\right]^2 + \lambda^{2}\left[\|\nabla\alpha\|^{2} + \|\nabla\beta\|^{2}\right]dx\, dy
\eeq \label{eq:newenergyf2}
This can be minimized by solving the associated Euler-Lagrange equations.
\begin{align*}
\pder[L]{\alpha} - \pder[]{x} \pder[L]{\alpha_x} - \pder[]{y} \pder[L]{\alpha_y} = 0 \\
\pder[L]{\beta} - \pder[]{x} \pder[L]{\beta_x} - \pder[]{y} \pder[L]{\beta_y} = 0
\end{align*}
where L is the integrand of the energy functional.
\beq
L = \left[I_{x}\alpha+I_{y}\beta+I_{z}\right]^{2} + \gamma^{2}\left[ D_x\alpha + D_y\beta + D_z\right]^2 + \lambda^{2}\left[\|\nabla\alpha\|^{2} + \|\nabla\beta\|^{2}\right]
\eeq

\begin{align*}
\pder[L]{\alpha} = 2 I_{x} \left( I_{x}\alpha + I_{y}\beta + I_{z} \right) + 2 \gamma^{2}D_{x} \left( D_{x}\alpha + D_{y}\beta + D_{z}\right) \\
\pder[L]{\beta} = 2 I_{y} \left( I_{x}\alpha + I_{y}\beta + I_{z} \right) + 2 \gamma^{2}D_{y} \left( D_{x}\alpha + D_{y}\beta + D_{z}\right) \\
\pder[]{x} \pder[L]{\alpha_x} = 2 \lambda^{2} \alpha_{xx} \\
\pder[]{y} \pder[L]{\alpha_y} = 2 \lambda^{2} \alpha_{yy} \\
\pder[]{x} \pder[L]{\beta_x} = 2 \lambda^{2} \beta_{xx} \\
\pder[]{y} \pder[L]{\beta_y} = 2 \lambda^{2} \beta_{yy} \\
2 I_{x} \left( I_{x}\alpha + I_{y}\beta + I_{z} \right) + 2 \gamma^{2}D_{x} \left( D_{x}\alpha + D_{y}\beta + D_{z}\right) - 2 \lambda^{2} \Delta \alpha = 0 \\
2 I_{y} \left( I_{x}\alpha + I_{y}\beta + I_{z} \right) + 2 \gamma^{2}D_{y} \left( D_{x}\alpha + D_{y}\beta + D_{z}\right) - 2 \lambda^{2} \Delta \beta = 0 
\end{align*}
After rearranging the terms, we get:
\begin{align*}
\left( I_{x}^2 + \gamma^{2}D_{x}^2 \right) \alpha + \left( I_{x}I_{y} + \gamma^{2}D_{x}D_{y} \right) \beta = \lambda^{2} \Delta \alpha - I_{x}I_{z} - \gamma^{2}D_{x}D_{z} \\
\left( I_{x}I_{y} + \gamma^{2}D_{x}D_{y} \right) \alpha + \left( I_{y}^2 + \gamma^{2}D_{y}^2 \right) \beta = \lambda^{2} \Delta \beta - I_{y}I_{z} - \gamma^{2}D_{y}D_{z}
\end{align*}
approximating the Laplacians of $\alpha$ and $\beta$,
\begin{align*}
\Delta \alpha \approx \rho \left( \overline{\alpha} - \alpha \right) \\
\Delta \beta \approx \rho \left( \overline{\beta} - \beta \right)
\end{align*}
where $\rho$ is a proportionality constant and, $\overline{\alpha}$ and $\overline{\beta}$ are local averages. These approximations are substituted for Laplacians and the terms in the equation are rearranged. 
\begin{align*}
\left( I_{x}^2 + \gamma^{2} D_{x}^2 + \lambda^{2} \right) \alpha + \left( I_{x}I_{y} + \gamma^{2}D_{x}D_{y} \right) \beta = \lambda^{2} \overline{\alpha} - \left( I_{x}I_{z} + \gamma^{2} D_{x}D_{z} \right) \\
\left( I_{x}I_{y} + \gamma^{2} D_{x}D_{y} \right) \alpha + \left( I_{y}^2 + \gamma^{2} D_{y}^2 + \lambda^{2} \right) \beta = \lambda^{2} \overline{\beta} - \left( I_{y}I_{z} +  \gamma^{2} D_{y}D_{z} \right)
\end{align*}
Determinants can be used to solve the above equations.
\begin{align*}
\alpha = \frac{Det_{\alpha}}{Det} \\
\beta = \frac{Det_{\beta}}{Det}
\end{align*}
\begin{align*}
Det 
&=
\begin{vmatrix}
\left( I_{x}^2 + \gamma^{2} D_{x}^2 + \lambda^{2} \right) & \left( I_{x}I_{y} + \gamma^{2}D_{x}D_{y} \right) \\
\left( I_{x}I_{y} + \gamma^{2} D_{x}D_{y} \right)         & \left( I_{y}^2 + \gamma^{2} D_{y}^2 + \lambda^{2} \right) \\
\end{vmatrix} \\
&=
\gamma^{2} \left( I_{x}D_{y} - I_{y}D{x} \right)^{2} + \lambda^{2} \left( \lambda^{2} + I_{x}^2 + I_{y}^2 + \gamma^{2}D_{x}^2 + \gamma^{2} D_{y}^2 \right)
\end{align*}
\begin{align*}
Det_{\alpha} 
&=
\begin{vmatrix}
\lambda^{2} \overline{\alpha} - \left( I_{x}I_{z} + \gamma^{2} D_{x}D_{z} \right) &  \left( I_{x}I_{y} + \gamma^{2}D_{x}D_{y} \right) \\
\lambda^{2} \overline{\beta} - \left( I_{y}I_{z} +  \gamma^{2} D_{y}D_{z} \right) &  \left( I_{y}^2 + \gamma^{2} D_{y}^2 + \lambda^{2} \right) \\
\end{vmatrix} \\
&=
\lambda^{2}\left( I_{y}^2 + \gamma^{2} D_{y}^2 + \lambda^{2} \right) \overline{\alpha} - \lambda^{2}\left( I_{x}I_{y} + \gamma^{2}D_{x}D_{y} \right) \overline{\beta} - \lambda^{2} \left( I_{x}I_{z} + \gamma^{2} D_{x}D_{z} \right) \\
& \quad - \gamma^{2} \left( I_{x}I_{z}D_{y}^2 + I_{y}^2 D_{x}D_{z} - I_{y}I_{z}D_{x}D_{y} - I_{x}I_{y}D_{y}D_{z} \right) \\
A 
&= \lambda^{2} \left( I_{x}I_{z} + \gamma^{2} D_{x}D_{z} \right) - \gamma^{2} \left( I_{x}I_{z}D_{y}^2 + I_{y}^2 D_{x}D_{z} - I_{y}I_{z}D_{x}D_{y} - I_{x}I_{y}D_{y}D_{z} \right) \\
Det_{\alpha} 
&= \lambda^{2}\left( I_{y}^2 + \gamma^{2} D_{y}^2 + \lambda^{2} \right) \overline{\alpha} - \lambda^{2}\left( I_{x}I_{y} + \gamma^{2}D_{x}D_{y} \right) \overline{\beta} - A \\
\end{align*}
\begin{align*}
Det_{\beta} 
&=
\begin{vmatrix}
\left( I_{x}^2 + \gamma^{2} D_{x}^2 + \lambda^{2} \right) &  \lambda^{2} \overline{\alpha} - \left( I_{x}I_{z} + \gamma^{2} D_{x}D_{z} \right) \\
\left( I_{x}I_{y} + \gamma^{2} D_{x}D_{y} \right)         &  \lambda^{2} \overline{\beta} - \left( I_{y}I_{z} +  \gamma^{2} D_{y}D_{z} \right) \\
\end{vmatrix} \\
&=
- \lambda^{2} \left( I_{x}I_{y} + \gamma^{2} D_{x}D_{y} \right) \overline{\alpha} + \lambda^{2} \left( I_{x}^2 + \gamma^{2} D_{x}^2 + \lambda^{2} \right) \overline{\beta} - \lambda^{2} \left( I_{y}I_{z} +  \gamma^{2} D_{y}D_{z} \right) \\
& \quad - \gamma^{2} \left( I_{y}I_{z}D_{x}^2 + I_{x}^2 D_{y}D_{z} - I_{x}I_{z}D_{x}D_{y} - I_{x}I_{y}D_{x}D_{z} \right) \\
B 
&= \lambda^{2} \left( I_{y}I_{z} +  \gamma^{2} D_{y}D_{z} \right) - \gamma^{2} \left( I_{y}I_{z}D_{x}^2 + I_{x}^2 D_{y}D_{z} - I_{x}I_{z}D_{x}D_{y} - I_{x}I_{y}D_{x}D_{z} \right) \\
Det_{\beta} 
&= - \lambda^{2} \left( I_{x}I_{y} + \gamma^{2} D_{x}D_{y} \right) \overline{\alpha} + \lambda^{2} \left( I_{x}^2 + \gamma^{2} D_{x}^2 + \lambda^{2} \right) \overline{\beta} - B \\
\end{align*}
\begin{align*}
Det \times \left(\alpha - \overline{\alpha} \right) = -\left[\gamma^{2}\left(I_{x}D_{y} - I_{y}D_{x}\right)^2 + \lambda^{2}\left(I_{x}^2 + \gamma^{2}D_{x}^2 \right) \right]\overline{\alpha} - \lambda^{2}\left(I_{x}I_{y} + \gamma^{2}D_{x}D_{y}\right)\overline{\beta} - A \\
Det \times \left(\beta - \overline{\beta}\right) = -\left[\gamma^{2}\left(I_{x}D_{y} - I_{y}D_{x}\right)^2 + \lambda^{2}\left(I_{y}^2 + \gamma^{2}D_{y}^2\right)\right]\overline{\beta} - \lambda^{2}\left(I_{x}I_{y} + \gamma^{2}D_{x}D_{y}\right)\overline{\alpha} - B \\
\end{align*}
\begin{align*}
\alpha^{n+1} = \overline{\alpha}^n - \frac{\left[\gamma^{2}\left(I_{x}D_{y}-I_{y}D_{x}\right)^2 + \lambda^{2}\left( I_{x}^2 + \gamma^{2}D_{x}^2\right)\right]\overline{\alpha}^n + \lambda^{2}\left(I_{x}I_{y} + \gamma^{2}D_{x}D_{y} \right)\overline{\beta}^n + A}{Det} \\
\beta^{n+1} = \overline{\beta}^n - \frac{\lambda^{2}\left(I_{x}I_{y} + \gamma^{2}D_{x}D_{y} \right)\overline{\alpha}^n + \left[\gamma^{2}\left(I_{x}D_{y}-I_{y}D_{x}\right)^2 + \lambda^{2}\left(I_{y}^2 + \gamma^{2}D_{y}^2\right) \right]\overline{\beta}^n + B}{Det} \\
\end{align*}

\clearpage
\bibliographystyle{IEEEbib}

\end{document}